\documentclass[eat,twocolumn]{jmlr}

\usepackage{rotating}% for sideways figures and tables
\usepackage{longtable}% for long tables
\usepackage{multirow}

\usepackage{booktabs}
\theorembodyfont{\upshape}
\theoremheaderfont{\scshape}
\theorempostheader{:}
\theoremsep{\newline}

\jmlrvolume{}
\firstpageno{1}
\jmlryear{2022}
\jmlrworkshop{Machine Learning for Health (ML4H) 2022}
\let\clearpage\relax
\title[EDEN]{EDEN : An Event DEtection Network for the annotation of Breast Cancer recurrences in administrative claims data}

\author{
\Name{Elise Dumas} \Email{elise.dumas@curie.fr}\\
\Name{Anne-Sophie Hamy} \Email{anne-sophie.hamy-petit@curie.fr}\\
\Name{Sophie Houzard} \Email{shouzard@institutcancer.fr}\\
\Name{Eva Hernandez} \Email{eva.hernandez@curie.fr}\\
\Name{Aullène Toussaint} \Email{aullene.toussaint@curie.fr}\\
\Name{Julien Guerin} \Email{julien.guerin@curie.fr}\\
\Name{Laetitia Chanas} \Email{laetitia.chanas@curie.fr}\\
\Name{Victoire de Castelbajac} \Email{victoiredecastelbajac@hotmail.fr}\\
\Name{Mathilde Saint-Ghislain} \Email{mathilde.saint-ghislain@curie.fr}\\
\Name{Beatriz Grandal} \Email{beatriz.grandalrejo@curie.fr}\\
\Name{Eric Daoud} \Email{daoudattoyaneric@gmail.com}\\
\Name{Fabien Reyal} \Email{fabien.reyal@curie.fr}\\
\Name{Chloé-Agathe Azencott} \Email{chloe-agathe.azencott@mines-paristech.fr}\\
\addr 26 Rue d'Ulm 75005 Paris
}

\begin{document}

\maketitle

\begin{abstract}
While the emergence of large administrative claims data provides opportunities for research, their use remains limited by the lack of
clinical annotations relevant to disease outcomes, such as recurrence in breast cancer (BC). Several challenges arise from the annotation of such endpoints in administrative claims, including the need to infer both the occurrence and the date of the recurrence, the right-censoring of data, or the importance of time intervals between medical visits. Deep learning approaches have been successfully used to label temporal medical sequences, but no method is currently able to handle simultaneously right-censoring and visit temporality to detect survival events in medical sequences. We propose EDEN (Event DEtection Network), a time-aware Long-Short-Term-Memory network for survival analyses, and its custom loss function. Our method outperforms several state-of-the-art approaches on real-world BC datasets. EDEN constitutes a powerful tool to annotate disease recurrence from administrative claims, thus paving the way for the massive use of such data in BC research.
\end{abstract}

\section{Introduction}
\label{sec:intro}

In 2020, 2.3 million individuals were newly diagnosed with breast cancer (BC) worldwide, accounting for one in four cancers in women \citep{sung_global_2021}. The natural history of BC is characterized by a long evolution, with two out of three deaths occurring more than five years after diagnosis \citep{early_breast_cancer_trialists_collaborative_group_ebctcg_relevance_2011}. Rather than overall mortality, the use of surrogate endpoints such as disease relapse is critical for evaluating the efficacy of treatment and reflect genuine oncologic outcomes of BC.

The emergence of large administrative data deriving from healthcare insurance claims provides opportunities for BC research, such as the measurement of BC financial burden \citep{luyendijk_assessment_2020}, the assessment of quality of care \citep{manca_electronic_2015}, the adherence to practice guidelines \citep{barretto_linking_2003}, or the identification of regional disparities \citep{salmeron_assessing_2021}.
However, beyond vital status, a major pitfall in medical and administrative data is the relative lack of well annotated proxys relevant to the outcomes of the disease. Hence, in BC, disease recurrence, either local or metastatic, is not routinely coded \citep{in_cancer_2014,warren_challenges_2015}, limiting the current
utility of BC administrative claims data. We hypothesized that BC relapse should theoretically be predictable from the registered medical examinations, and could serve as a useful clinical annotation.

Deep learning methods were successfully used on longitudinal event sequences \citep{xiao_opportunities_2018} to embed medical concepts \citep{jia_spatio-temporal_2017, peng_bitenet_2020}, to diagnose medical conditions \citep{baumel_multi-label_2018,kam_learning_2017,lipton_learning_2017, fouladvand_identifying_2022}, or to  predict early clinical events \citep{choi_doctor_2016,zhang_leap_2017, zeng_pretrained_2022}. So far, the classification of time to survival event (\textit{e.g.} BC recurrence) from longitudinal data (\textit{e.g.} administrative claims data) received little attention in the machine learning literature (see Appendix \ref{apd:related_work} for a review). Survival outcome labeling poses several constraints compared with typical deep learning applications. Indeed, (1) survival functions exhibit a particular shape, with the probability that the event already occurred increasing over time; (2) most datasets used for training are right-censored, with a high proportion of patients never experiencing the event; and (3) the intervals between time points of a patient’s care pathway are critical and should be considered in the model, since treatments are presumed to be part of the initial therapy of the disease if administered within a short period of time after diagnosis, while they suggest a recurrence if administered long after diagnosis.

In this work, we propose EDEN (Event DEtection Network), a non-parametric bi-directional LSTM network which aims at detecting survival events in longitudinal, irregularly time-spaced and right-censored medical sequences, and its custom loss function. EDEN is an extension of T-LSTM \citep{baytas_patient_2017} that uses the chain rule derivation of discretized hazards to model the event rate function. We apply EDEN to administrative claim datasets of BC patients to identify the date and the type of BC relapses. We compare the results with several previously published methods and validate them on an independent dataset.

\section{Methods}
\label{sec:methods}

\begin{figure}
\floatconts
  {fig:example_data}
  {\vspace*{-10mm} \caption{Example of the medical record of patient $i$ who experienced the event $s$ (typically BC locoregional relapse) at the time of his/her fourth medical visit.}}
  {\includegraphics[width=0.5\textwidth]{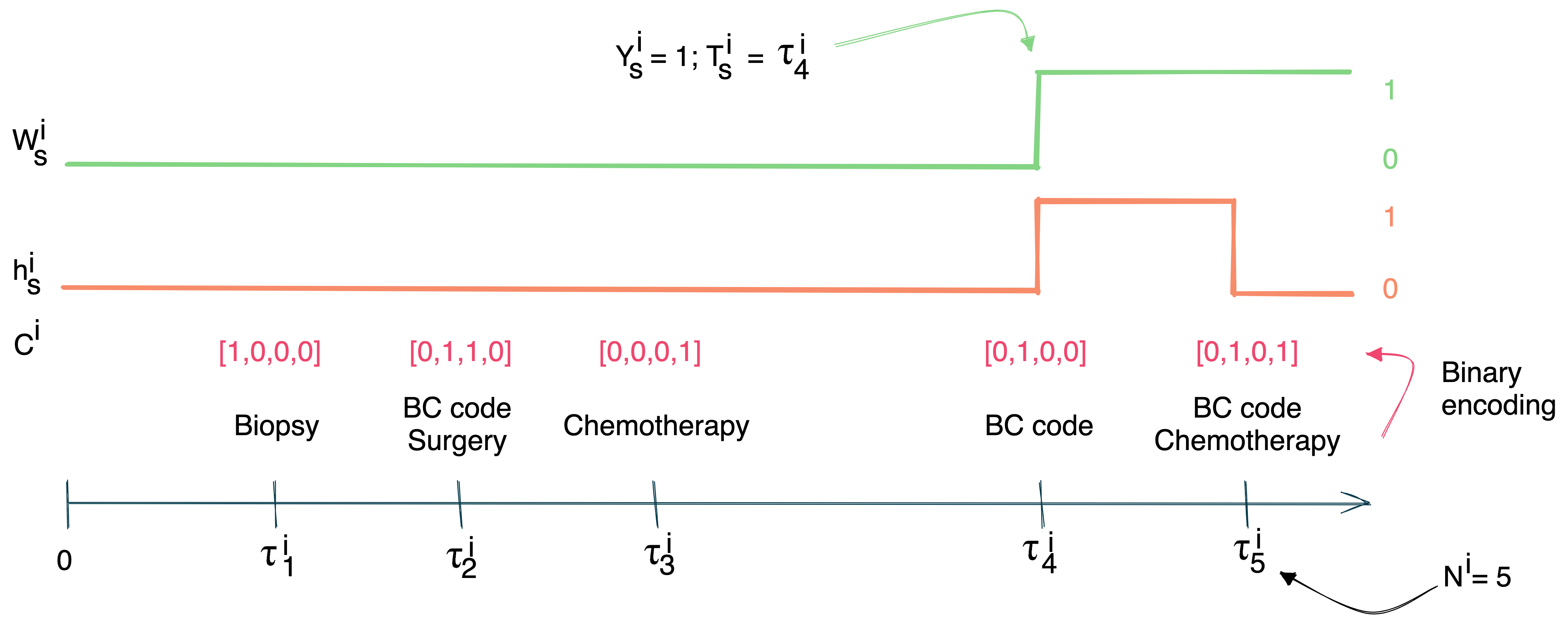}}
\end{figure}

\subsection{Notations}
\label{subsec:methods_notations}

We assume our dataset is a collection of patient records comprising:

\paragraph{(i) a time-labeled sequence of administrative medical codes derived from healthcare claims  $X$:}
For a given patient $i$, $X^i$ consists of a sequence of $N^i$ medical visits $M_1^i …, M_{N^i}^i$, ordered by time. $X^i$ contains all medical visits until the end of the study, potentially even those occurring after the event of interest. The $j$-th medical visit of patient $i$, denoted by $M_j^i $, is a pair $M_j^i = (C_j^i,\tau_j^i)$ where :
\begin{description}
    \item $C_j^i \in \{0,1\}^p$ is a binary vector representing the medical codes recorded during the medical visit $j$. At index $k$, $C_{j,k}^i=1$ if the medical code $k$ was recorded during visit $j$, and $C_{j,k}^i=0$ otherwise. $p \in \mathbb{N}^*$ is the size of the medical code vocabulary.
    \item $\tau_j^i \in \mathbb{R^+}$ is the time elapsed between the beginning of the study and the date of visit $j$. 
\end{description}

\paragraph{(ii) Times to event $T$ and (iii) event indicators $Y$}
We consider $S$ possible events (survival endpoints), which can occur successively within a patient trajectory. For example, in our experiments, there are $S=3$ possible events : metastatic relapse, locoregional relapse, and second cancer from another localization. If the event $s$ is observed before the end of the study, $Y_s=1$, and the time to event $T_s$ is set to the time elapsed between the onset of the study and the date of the event. Otherwise, if the event $s$ is not observed before the end of the study, $Y_s=0$, and $T_s$ is set to the time elapsed between the onset of the study and the date of last news with the patient. In that case, the patient is said to be right-censored.

\subsection{Problem modeling}
\label{subsec:methods_modeling}

Our goal is to learn simultaneously the event indicators $Y_s$ and the times to event $T_s$, by inferring a discretized version of the patient event rate function $W_s^i(t) = Pr(T_s^i \leq t)$, for $t \in \tau_1^i,…, \tau_{N^i}^i$. For right-censored patients, $W_s^i(t)=0$ at all time stamps $\tau_j^i$. For patients experiencing the event $s$, $W_s^i(t)=0$ before the event ($t< T^i$) and $W_s^i(t)=1$ after the event ($t \geq T^i$). In that scope, $Y_s$ and $T_s$ can be respectively approached by: 
\[\tilde{Y_s^i} = \max_{1 \leq r \leq N^i}W_s^i(\tau_{r}^i)\]
\[\tilde{T_s^i} = \tau_{\mu}^i \text{ where } \mu = \min \{r | W_s^i(\tau_{r}^i) = 1\}\]

In our model, $W_s^i$ is learned through the discretized version of the instantaneous hazard rate $h_s^i(t)$, which models the instantaneous occurrence probability of the event $s$ at time $t$ given that the event has not occurred before.
Formally, $$h_s^i(\tau_r^i) = Pr\big(T_s^i \in \big[\tau_{r-1}^i,\tau_{r}^i\big] \big| T_s^i > \tau_{r-1}^i\big)$$
$W_s^i$ can be directly derived from $h_s^i$ (see Appendix \ref{apd:survival_notation} for demonstration):
\begin{equation}\label{eq:WprodH}
W_s^i(\tau_{j}^i) =  1 - \prod_{r \leq j} \big( 1 - h_s^i(\tau_r^i)\big)
\end{equation}

\subsection{Model}

Figure \ref{fig:eden_model_overview} shows the high-level overview of the proposed model. The goal is to learn the event rate function $W$ at each time stamp. We first embed the medical codes in a continuous vector space of dimension $n\textsubscript{emb}$: \[X^i_\textsubscript{emb} = M_\textsubscript{emb}C^{i}\]
with $M\textsubscript{emb} \in \mathbb{R}^{n\textsubscript{emb}\times p}$ being a network parameter optimized by the model in an end-to-end approach. The embedding size $n\textsubscript{emb}$ is a model hyper-parameter tuned by random search (see Appendix \ref{apd:hyperparameter}).

\begin{figure*}
\floatconts
  {fig:eden_model_overview}
  {\vspace*{-8mm} \caption{The proposed EDEN model}}
  {\includegraphics[width=0.6\textwidth]{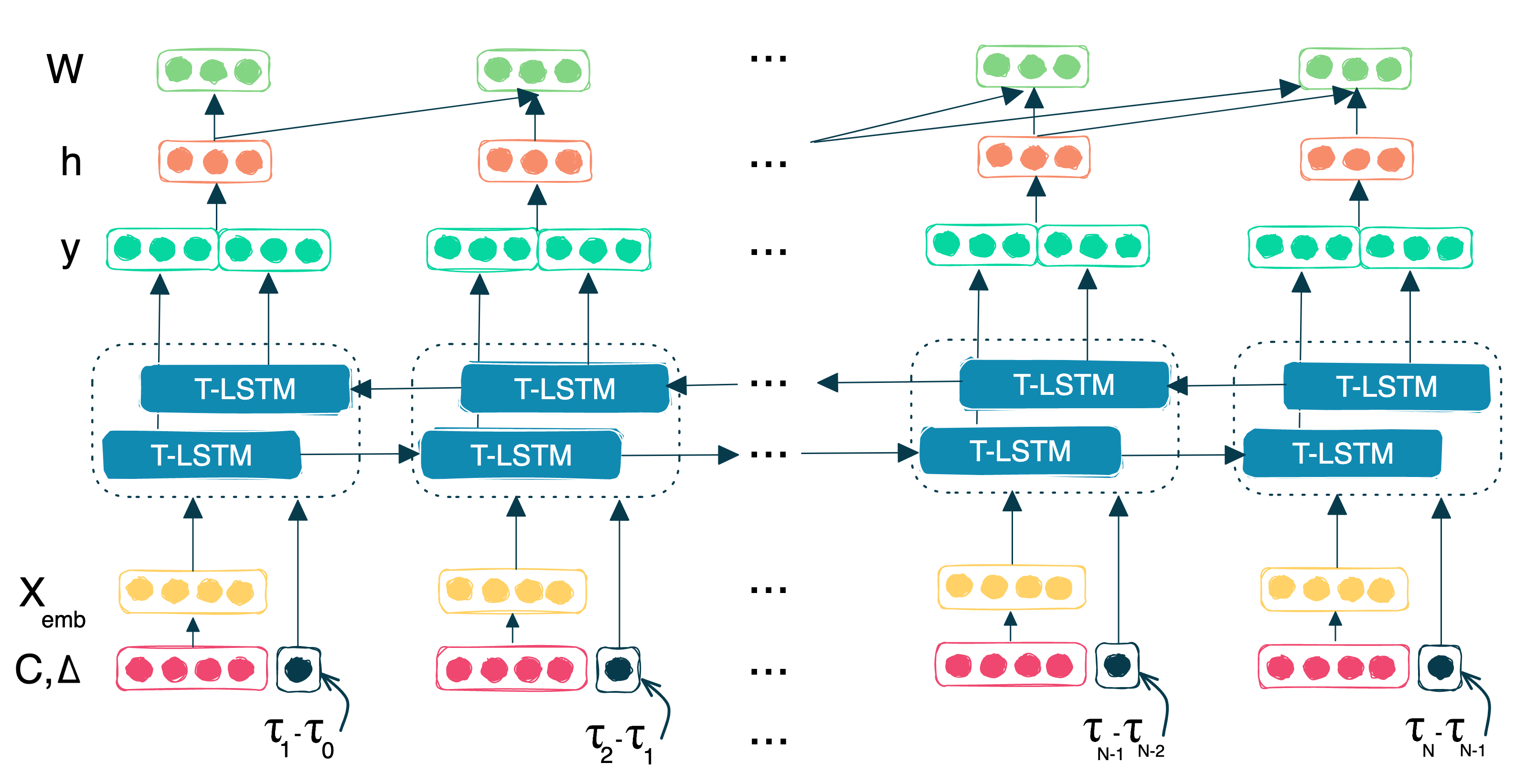}}
\end{figure*}

We expanded the Time-Aware LSTM (T-LSTM) network introduced by \citet{baytas_patient_2017}, into a bi-directional network, as explained in Appendix \ref{apd:bi-LSTM}. The embedded medical visits, along with the interval between events, are used as input to this many-to-many bi-directional T-LSTM network:
\begin{align*}
y^i_1,...,y^i_{N^i} = \text{Bi-T-LSTM}(&(X_{emb,1}^i,\Delta_1^i), ..., \\
& (X_{emb,N^i}^i, \Delta_{N^i}^i))
\end{align*}
The so-computed outputs are fed into a fully connected layer to predict the instantaneous hazard rates $\tilde{h}$: $\tilde{h}^i_1, ... \tilde{h}^i_{N^i} = \text{sigmoid}\text{(FC(}(y^i_1,...,y^i_{N^i}))$. Finally, following an approach proposed elsewhere \citep{ren_deep_2018,gensheimer_scalable_2019}, $\tilde{h}$ is converted into the inferred event rate function $\tilde{W}$ using (\ref{eq:WprodH}).

In this way, the prediction of $\tilde{W}^i_s$ at time $\tau_{r}^i$ is connected to all the previous model outputs $\tilde{h}^i_s(\tau_{1}^i),...\tilde{h}^i_s(\tau_{r}^i)$. This chain rule calculation improves the propagation of the gradient across the whole sequence, preventing the gradient signal from being too sparse outside of the true event time $T^i_s$. It also ensures that the event rate function $W^i_s$ is non-decreasing.

\subsection{Loss function}

EDEN is trained by minimizing a custom loss function $\mathcal{L}_{Total}$. $\mathcal{L}_{Total}$ accounts for the specific form of the event rate function $W$, and is specifically designed to handle right-censored data. It can be decomposed into four terms $\mathcal{L}_{Total} = \alpha_1\mathcal{L}_{1}+\alpha_2\mathcal{L}_{2}+\alpha_3\mathcal{L}_{3}+\alpha_4\mathcal{L}_{4}$, where $\alpha_1, ..., \alpha_4 \in \mathbb{R^+}$ are model hyper-parameters.

$\mathcal{L}_{1}$ is the binary-cross entropy (BCE) between $\tilde{W}$ and $W$, weighted to account for the over-representation of $0$ over $1$ in the event rate functions $W$. It is the standard sequence-to-sequence loss function \citep{goodfellow_deep_2016}, which exhibits good smoothness properties and eases model convergence, but did not account for the survival shape of $W$, nor for the presence of right-censored units. $\mathcal{L}_{2}$ and $\mathcal{L}_{3}$ control for the error over the uncensored samples. $\mathcal{L}_{2}$ penalizes low values of $\tilde{W}$ at the exact time of relapse. $\mathcal{L}_{3}$ penalizes high values of $\tilde{W}$ at the time of the last medical visit before relapse. $\mathcal{L}_{4}$ controls for the error over the censored samples, by penalizing high values of $\tilde{W}$ at the time of the last medical visit of the sequence. Turning $\mathcal{L}_{2}$, $\mathcal{L}_{3}$ and $\mathcal{L}_{4}$ altogether to zero is sufficient to learn the exact event rate function $W$. Indeed, it follows from the chain rule derivation (Equation 1) that $W$ is fully characterized by the shift from zero to one at the date of the event for uncensored units and by the last value being zero for censored units. Details on the loss function is given is Appendix \ref{apd:loss}.

\section{Experiments}
\label{sec:data_preprocessing}

We implemented EDEN on two French claims datasets structured from the National Health Data System (SNDS) database \citep{tuppin_value_2017}. We used two independent cohorts (Appendix \ref{apd:snds_data}). Cohort A included 5,892 patients treated for a primary BC in the Institut Curie between 2009 and 2012 and over 45 years of age at diagnosis. We split Cohort A into training, validation and test sets with ratio 3:1:1. Cohort B included 800 patients treated with neoadjuvant chemotherapy for a primary invasive BC at the Institut Curie between 2009 and 2017. We used Cohort B as a fully independent test set. We annotated manually BC relapses from in-house medical records for both cohorts. We considered three types of events: locoregional relapse (local invasive ipsilateral recurrence, regional recurrence, or invasive contralateral BC), distant metastatic relapse, and second cancer (second primary invasive cancer, excluding breast) \citep{hudis_proposal_2007}. Dataset statistics are provided in Appendix \ref{apd:data_stat}.

We compared EDEN performance to several other models as baselines : ad-hoc decision rules \citep{izci_systematic_2020}, LSTM, Dipole \citep{ma_dipole_2017}, and Timeline \citep{bai_interpretable_2018}. Apart from Timeline, baseline models did not account for the irregular time interval between medical visits. Implementation details for EDEN and baselines are provided in Appendices \ref{apd:hyperparameter} and \ref{apd:baseline}. EDEN code is available online (\href{https://github.com/rt2lab/eden}{github.com/rt2lab/eden}).

We used several metrics to measure the model accuracy for each type of event: (i) the area under the receiving operating characteristic curve, denoted by AUC; (ii) the F1-score, denoted by F1; (iii) the accuracy, denoted by Acc; (iv) the integrated Brier score \citep{graf_assessment_1999}, denoted by Brier; (v) the concordance index \citep{brentnall_use_2018}, denoted by C; and (vi) the mean interval between the true and the predicted event date in days, denoted by $\Delta_T$.

\section{Results}
\label{sec:results}

\paragraph{Comparison with baselines}

%Table 1 AUC Acc et delta t
%Version number 2
\setlength{\tabcolsep}{1.2pt}
\begin{table*}
\tiny
\floatconts
{tab:results_performance}
{\caption{Performances comparison between models. The best scores are presented in bold.\vspace*{-5mm}}}
  {\begin{tabular}{|c|c|c|c|c|c|c|c||c|c|c|c|c|c||c|c|c|c|c|c|}
  \toprule
  & & \multicolumn{6}{c||}{\color[HTML]{f4a261} Locoregional} & \multicolumn{6}{c||}{\color[HTML]{e76f51} Metastatic} & \multicolumn{6}{c|}{\color[HTML]{DFAC2A} $2^{nd}$ cancer}\\
  \midrule
  Cohort & Model & AUC & Acc & $\Delta_T$ & F1 & Brier & C & AUC & Acc & $\Delta_T$ & F1 & Brier & C & AUC & Acc & $\Delta_T$ & F1 & Brier & C\\
  \midrule
  \multirow{5}{*}{A} & \bfseries Ad-hoc & 85.8 & 96.3 & 59 & 69.1 & \bfseries 0.017 & 0.229 & 94.2 & 96.5 & -51 & 76.3 & 96.5 & 0.112 & 95.6 & 93.6 & \bfseries 7 & 54.5 & 0.027 & 0.130\\
  
  & \bfseries LSTM & 94.7 & 95.8 & 41.4 & 67.6 & 0.055 & 0.162 & 96.4 & 97.3 & 28.1 &
  79.6 & 0.037 &  0.082 & 98.0 & 97.7 & 46.6 & 73.9 & 0.035 & 0.143\\
  
  &\bfseries Dipole & 92.0 & 94.8 & 239.8 & 39.2 & 0.126 & 0.126 &
  96.9 & 97.9 & \bfseries 0.6 & 81.1 & 0.026 & 0.048  & 97.6 & 97.6 & 188.4 & 58.6 & 0.059 & \bfseries 0.047\\
  &\bfseries Timeline & 95.0 & 96.1 & 16.8 & 68.7 & 0.022 & \bfseries{0.112} & \bfseries 98.5 & 97.9 & 3.6 & 83.0 & 0.010 & \bfseries 0.044 & 95.4 & 97.3 & 58.0 & 66.1 & 0.023 & 0.122\\
  & \bfseries EDEN & \bfseries 96.9 & \bfseries 97.2 & \bfseries 5.3 & \bfseries 73.8 & 0.019 & 0.127 & 
  98.4 & \bfseries 98.5 & -1.9 & \bfseries 87.9 & \bfseries 0.008 & 0.047 &
  \bfseries 98.7 & \bfseries 98.3 & 44.5 & \bfseries 76.0 & \bfseries 0.021 & 0.056\\

  \midrule
  \multirow{5}{*}{B} & \bfseries Ad-hoc & 75.3 & 95.3 & -18 & 40.6 & 0.071 & 0.303 &
  98.7 & 97.5 & -14 & 83.6 & 0.028 & 0.043 & 90.7 & 98.0 &  -56 & 38.5 & 0.022 & 0.168\\
  
  & \bfseries LSTM & 91.7 & 95.8 & 35.4 &  47.0 & 0.098 & 0.200 &
  98.9 & 97.3 & 28.0 & 80.9 & 0.059 & 0.062 & 90.5 & 99.1 & 13 & 47.4 & 0.043 & 0.183\\
  
  &\bfseries Dipole & \bfseries 95.4 & \bfseries 96.8 & 407 & 27.8 & 0.148 & \bfseries 0.084 &
  99.2 & 98.2 & 42 & 84.4 & 0.038 & 0.039 &
  \bfseries 92.5 & 99.3 & -75.4 &  28.9 & 0.045 & \bfseries 0.141\\
  
  &\bfseries Timeline & 93.5 & 95.2 & 18.6 & 43.2 & 0.070 & 0.147 & 99.0 & 98.1 & 9.5 & 86.1 & 0.027 & 0.055 &
  89.8 & 99.2 & -17.9 & 50.4 & 0.019 & 0.214\\
  
  & \bfseries EDEN & 92.4 & 96.3 & \bfseries 8.8 & \bfseries 49.4 & \bfseries 0.053 & 0.149 &
  \bfseries 99.3 & \bfseries 98.3 & \bfseries 1.1 & \bfseries 87.2 & \bfseries 0.017 & \bfseries 0.037 & 89.6 & \bfseries 99.4 & \bfseries 7.1 & \bfseries 58.2 & \bfseries 0.018 & 0.193 \\
  
  \bottomrule
  \end{tabular}}
\end{table*}

Results for all models are presented in Table \ref{tab:results_performance}. EDEN was the best or second best performing model for most metrics on cohort A, and showed high performance both in the detection of breast cancer relapses (AUC, accuracy, and F1) and in the inference of the dates of relapse ($\Delta_T$, Brier and C).

EDEN outperformed from far all other models for locoregional relapses on cohort A (F1-score of 73.8\% \textit{versus} 68.7\% for Timeline). Detecting this type of recurrence was particularly challenging because the treatments of locoregional relapses are identical to the treatments of the initial disease, while distant metastasis and second cancer are often associated with specific diagnosis codes or molecules. EDEN also achieved good performance on cohort A for metastatic relapses, with an F1-score of 87.9\%; and for second cancer relapse, with an F1-score of 76.0\%. EDEN dated the relapses with high accuracy, with an averaged error of respectively 5.3 and -1.9 days for locoregional and metastatic recurrences on cohort A.

%No model clearly outperforms the others for the detection of second cancer on cohort A. Indeed, the annotation of second cancer relapse from clinical in-house medical records was impaired by the poor quality of the follow-up except for BC disease: while relapses from breast cancer were mostly treated in-house, cancers of other localization were likely to be treated in other care centers.

EDEN was the best or the second best performing model for most metrics on the independent test set (cohort B), despite substantial variations in patients characteristics, with patients from cohort B younger and with a more advanced and aggressive disease than in cohort A. This performance illustrates the robustness of EDEN to the study population. Overall, the results suggest that EDEN was able to robustly decipher complex and irregularly time-spaced longitudinal medical sequences to detect both the occurrence and the date of BC relapses.

\paragraph{Network learning dynamics and ablation study}

EDEN convergence is studied in Appendix \ref{apd:loss_curves}. We also conducted an ablation study to determine the extent to which the different components of EDEN and of its loss function improved performance. Results are detailed in Appendix \ref{apd:ablation}. They suggest that EDEN's components contributed all together to the performance of the model, with the bi-directional architecture and the custom loss function leading to the highest accuracy gaps compared to the standard T-LSTM. Interestingly, the integration of the chain rule derivation of the event rate function from the instantaneous hazard tended to decrease the performance of the model when trained with BCE only, highlighting the close relationship between the survival output and the survival components of the loss ($\mathcal{L}_2$, $\mathcal{L}_3$ and $\mathcal{L}_4$). 

\paragraph{Interpretation of results}
We further evaluated the interpretability of EDEN by observing the importance of each medical code on the network output. The results presented in Appendix \ref{apd:results_interp} were in accordance with clinical practices,  with codes of molecules approved only in the metastatic setting associated with an increase in the event rate function for metastatic relapse. The recording of local BC diagnostic (breast biopsy and cytology) or surgical procedures (lumpectomy) was associated with an increase in the event rate function of locoregional relapse. Kaplan-Meier survival curves are also presented in Appendix \ref{apd:survival_curves}.

\section{Discussion}
\label{sec:discussion}

In this paper, we propose EDEN (Event DEtecton Network), a deep learning recurrent neural network able to detect and classify survival endpoints in longitudinal data with irregular elapsed times between the successive events. EDEN benefits from a custom loss function developed for survival analyses. Based on EDEN, we annotated breast cancer relapses on administrative claim datasets. EDEN outperformed several other methods, including clinical-based decision rules and state-of-the-art deep learning methods.

Although handling of time irregularities may hurt robustness to new settings \citep{javidi_identification_2022}, EDEN showed robust performance on the independent test set. Another pitfall of neural networks is that they may fail to adapt to changes in medical practices over time \citep{guo_evaluation_2022}. We tried to mitigate this issue in EDEN by aggregating the raw medical codes into clinically relevant categories, as advised elsewhere \citep{nestor_feature_2019}, but we cannot fully rule out that clinical shift may hurt the performance of EDEN over time or across hospitals.

To conclude, EDEN is a powerful tool to annotate disease recurrence from complex longitudinal data. Our method enables to structure relevant proxies for disease outcomes, thus paving the way for the massive use of administrative claims in breast cancer research.

\acks{We thank Monoprix* for funding the study; and the Ecole polytechnique, for providing Elise Dumas with a PhD grant (AMX). We also thank Thomas Walter and Adeline Fermanian for the careful proofreading of the manuscript.
}

%\newpage
\bibliography{biblio_eden_v5}

\newpage
\onecolumn
\appendix

\section{Related work}
\label{apd:related_work}

\paragraph{Annotating BC recurrence from administrative claims}

Several algorithms to identify BC recurrences from administrative claims data have been proposed \citep{izci_systematic_2020}. They rely on decision rules driven by clinical practices \citep{hassett_validating_2014,rasmussen_validated_2021} or on classification and regression trees (CART) derived from hand-crafted features \citep{chubak_administrative_2012,nordstrom_identification_2012,xu_development_2019}. Most of them were both calibrated and tested on the same dataset \citep{mcclish_using_2003,sathiakumar_accuracy_2017}. Some studies focused only on distant (metastatic) recurrences \citep{chawla_limited_2014,whyte_evaluation_2015}, while others 
did not distinguish the type of BC relapse when identifying an event \citep{chubak_electronic_2017,kroenke_enhancing_2016}. Most studies did not infer the date of recurrence, which is the cornerstone of survival analysis. So far, no algorithm simultaneously identifies (i) the occurrence, (ii) the date, and (iii) the type (locoregional, metastatic, or second primary tumor from another localization) of BC recurrence.

\paragraph{Learning from longitudinal medical data}

Deep learning methods recently achieved great progress in labeling temporal medical sequences \citep{xiao_opportunities_2018}. Recurrent Neural Networks (RNNs) \citep{rumelhart_learning_1986} and their gated extensions such as Long-Short-Term Memory (LSTM) \citep{gers_learning_1999} outperformed hand-crafted decision rules and classical machine learning tools in the early prediction of several conditions such as sepsis \citep{kam_learning_2017,lin_early_2018} or Alzheimer’s’ disease \citep{hong_predicting_2019,li_diagnosis_2022}. Bi-directional RNN and LSTM \citep{schuster_bidirectional_1997,graves_framewise_2005} scan sequences in both forward and backward directions, using all available input information in the past and future, improving the performance of unidirectional models with respect to several medical applications \citep{jagannatha_bidirectional_2016,ma_dipole_2017,zhu_predicting_2019,peng_bitenet_2020}.
Temporal convolutional networks \citep{jarrett_dynamic_2020,catling_temporal_2019} and transformer frameworks \citep{fouladvand_identifying_2022,zeng_pretrained_2022} were also proposed as alternative deep learning methods to capture longitudinal effects from medical records.

\paragraph{Handling time irregularity}

Irregular timing between events are critical in sequence labeling for medical applications. 
Several works propose to extended classical recurrent networks to handle time irregularity \citep{weerakody_review_2021}.
In \citet{pham_deepcare_2016,choi_retain_2017,lipton_learning_2017,kabeshova_zimm_2020, peng_bitenet_2020}, raw or embedded time intervals are concatenated with the current input at each time step.  Standard methods for continuous time series consist in dividing the timeline into fixed-width intervals and impute missing observations by average or exponential decay \citep{lipton_learning_2017,che_recurrent_2018}, but they are not adequate for sequences of discrete entries. Other authors amend the network architecture \citep{baytas_patient_2017,ye_lsan_2020,bai_interpretable_2018, mozer_discrete_2017-1, xiao_modeling_2017}. Among them, T-LSTM \citep{baytas_patient_2017} decomposes the memory of the previous time steps into discounted short-term memory and long-term memory, while Timeline \citep{bai_interpretable_2018} weights medical codes and visits using data-driven and time-dependent functions. Neural Ordinary Differential Equations were also recently proposed as networks replacing RNN discrete state transitions by continuous dynamics \citep{chen_neural_2019, rubanova_latent_2019}.

\paragraph{Learning from survival right-censored data}

Survival (or time-to-event) analyses arise when interest is focused on the time elapsing from the beginning of the study until an event occurs \citep{bewick_statistics_2004}. The event of interest may not be observed for all patients (who may be lost to follow-up before experiencing it), frequently resulting in right-censored data. In traditional statistics, Kaplan-Meier models estimate non-parametric survival curves \citep{kaplan_nonparametric_1958}; while Cox proportional hazard (Cox-PH) is a semi-parametric method modeling hazard through time \citep{cox_regression_1972}. Other classical methods rely on statistical distributions such as Gompertz \citep{wilson_analysis_1994} or Weibull \citep{anderson_nonproportional_1991} to model times to event. Traditional survival methods were embedded into deep learning methods by \cite{ching_cox-nnet_2018,katzman_deepsurv_2018, nagpal_deep_2021-2, wang_survtrace_2022, lee_deephit_2018}.

The aforementioned methods are static, as opposed to dynamic, because they cannot handle longitudinal data. Traditional dynamic survival methods include (i) landmarking which consists in building repeated static survival models at pre-defined times \citep{van_houwelingen_dynamic_2007}, and (ii) joint modeling, which jointly learns the distributions of the longitudinal process and of the times to event \citep{hickey_joint_2016}.
Several approaches extended joint modeling to deep learning methods \citep{nagpal_deep_2021-1,sun_attention-based_2021}. While several of them can handle irregular time stamps \citep{jeanselme_deepjoint_2022,lee_dynamic-deephit_2020,jarrett_dynamic_2020}, they infer the time to event at the end of the sequence only, so that they can only be used to predict future events and are not adequate for the annotation of survival endpoints in medical visit sequences, where the event of interest may occur during the input sequence.

Other authors proposed sequence-to-sequence network with predictions at each time stamp. Among them, several methods modeled the time to event at each time stamp by a positive survival statistical distribution, thus constraining that the event did not occur before the end of the sequence \citep{putzel_dynamic_2021,martinsson_model_2017}. Non-parametric online methods consisting in inferring a discretized version of the survival function by chain rule multiplication of hazards were also proposed \citep{zhang_survival_2020, ren_deep_2018, vale-silva_long-term_2021}. They enable to handle events occurring during the input sequence, and demonstrated good convergence since the chained derivation of the survival function may benefit to the propagation of the gradient across the whole sequence \citep{ren_deep_2018}.

So far, no survival model encompassing time irregularity is directly applicable to sequence event detection. In this paper, we propose EDEN (Event DEtection Network), a non-parametric bi-directional LSTM network which aims at detecting survival events in longitudinal, irregularly time-spaced and right-censored medical sequences. EDEN is an extension of T-LSTM \citep{baytas_patient_2017}, which uses the chain rule derivation of discretized hazards to model the event rate function.

\section{Relation between event rate and hazard rate functions}\label{apd:survival_notation}

Using the notations introduced in Sections \ref{subsec:methods_notations} and \ref{subsec:methods_modeling}, we have : 
\[W_s^i(\tau_r^i) = Pr(T_s^i \leq \tau_r^i) = 1 - Pr(T_s^i > \tau_r^i).\]

We denote by $\Delta_k^i=\big[\tau_{k-1}^i,\tau_{k}^i\big]$, and $\tau_{0}^i = 0$.  It follows that: 
\begin{align*}
Pr(T_s^i > \tau_r^i) &= Pr(T_s^i \notin \Delta_1^i, ...,T_s^i \notin \Delta_r^i)\\
%&= Pr(T_s^i \notin \Delta_1^i)\cdot Pr(T_s^i \notin \Delta_2^i | T_s^i \notin \Delta_1^i)\cdot ... \cdot Pr(T_s^i \notin \Delta_r^i | T_s^i \notin \Delta_1^i,...T_s^i \notin \Delta_{r-1}^i)\\
&= \prod_{1 \leq k \leq r}Pr(T_s^i \notin \Delta_k^i | T_s^i \notin \Delta_1^i,...,T_s^i \notin \Delta_{k-1}^i) \\
&= \prod_{1 \leq k \leq r} \Bigl[ 1 - Pr(T_s^i \in \Delta_k^i | T_s^i \notin \Delta_1^i,...,T_s^i \notin \Delta_{k-1}^i) \Bigr] \\
& = \prod_{1 \leq k \leq r} \Bigl( 1 - h_s^i(\tau_k^i) \Bigr).
\end{align*}

Finally, 

\[ \boxed{W_s^i(\tau_r^i) = 1 - \prod_{1 \leq k \leq r} \Bigl( 1 - h_s^i(\tau_k^i) \Bigr).}\]

\section{Time-Aware LSTM networks (T-LSTM)}\label{apd:bi-LSTM}

Time-Aware LSTM networks, denoted ad T-LSTM, are recurrent neural networks able to handle irregular time intervals in longitudinal medical data. They were introduced by \citet{baytas_patient_2017} in 2017. In the T-LSTM architecture, at each time step $t$, the current memory $C_{t-1}$ is divided into short-term memory $C_{t-1}^S$ and long-term memory $C_{t-1}^L$:

\[C_{t-1}^S = \text{tanh}(W_dC_{t-1}+b_d) \text{ with } W_d\in\mathbb{R}^{n\textsubscript{hidden}\times n\textsubscript{hidden}}, b_d\in \mathbb{R}^{n\textsubscript{hidden}}\]
\[C_{t-1}^L= C_{t-1} - C_{t-1}^S.\]

Short-term memory is altered using a discount function $g$ taking as input the time interval with the previous medical visit $\Delta_t$.
\[\hat{C}_{t-1}^S = C_{t-1}^S * g(\Delta_t).\]

The authors proposed several discount functions. In this paper, we use the function $g$ defined as:
\[ g(\Delta_t) = \frac{1}{\text{log}(e+\Delta_t)}.\]

The adjusted previous memory is the sum of the long-term memory and the altered short-term memory.

\[C_{t-1}^{*} = C_{t-1}^L + \hat{C}_{t-1}^S.\]

The adjusted previous memory $C_{t-1}^{*}$ is used as candidate memory to a standard Long-Term Memory (LSTM) unit:

\begin{align*}
\tag{Forget gate}
f_t &= \sigma(W_fx_t+U_fh_{t-1}+b_f)\\
\tag{Input gate}
i_t &= \sigma(W_ix_t+U_ih_{t-1}+b_i)\\
\tag{Output gate}
o_t &= \sigma(W_ox_t+U_oh_{t-1}+b_o)\\
\tag{Candidate memory}
\tilde{C} &= \text{tanh}(W_cx_t+U_ch_{t-1}+b_c)\\
\tag{Current memory}
g_t &= f_t*C_{t-1}^{*} + i_t*\tilde{C}\\
\tag{Current hidden state}
y_t &= o_t * \text{tanh}(g_t).\\
\end{align*}

with $W_f,W_i,W_o,W_c \in \mathbb{R}^{n\textsubscript{emb}\times n\textsubscript{hidden}}$,$U_f,U_i,U_o,U_c \in \mathbb{R}^{n\textsubscript{hidden}\times n\textsubscript{hidden}}$, and $b_f,b_i,b_o,b_c \in \mathbb{R}^{n\textsubscript{hidden}}$.

In this work, we used T-LSTM units as the basic module of a bi-directional recurrent network. We compute forward and backward hidden states by passing the input into a forward T-LSTM and a backward T-LSTM network respectively. Next, we concatenate the forward and backward hidden states, to obtain the final hidden state $y^i$. The quantity $y^i$ is used as input to the output fully connected layer.

\section{Survival loss function}\label{apd:loss}

In this section, we detail the four components of the survival loss function. They were inspired from other deep learning methods focusing on survival data \citep{lee_deephit_2018,ren_deep_2018}.

$\mathcal{L}_{1}$ is the binary-cross entropy (BCE) between $\tilde{W}$ and $W$, weighted to account for the over representation of $0$ over $1$ in the event rate functions $W$:
\begin{align*}
 \mathcal{L}_{1}(W^i, \tilde{W}^i) =  \frac{1}{N} \sum_{1 \leq i \leq N} \frac{1}{N^i} \sum_{1 \leq k \leq N^i} \frac{1}{S} \sum_{1 \leq s \leq S} & \Bigl( \beta_1 \times W^{i}_s(\tau^{i}_k) \times \text{log}(\tilde{W}^{i}_s(\tau^{i}_k))\\
 & + \beta_0 \times (1-W^{i}_s(\tau^{i}_k)) \times \text{log}(1-\tilde{W}^{i}_s(\tau^{i}_k)) \Bigl),
\end{align*}
where $N$ is the number of patients and $\beta_0, \beta_1 \in \mathbb{R}$ are balancing weights computed as: 
\begin{align*}
    \tilde{\beta_1} &= \frac{\text{\# medical visits}}{\text{\# medical visits occurring after events}}= \frac{S \times \sum_{1 \leq i \leq N}N^i}{\sum_{1 \leq i \leq N}\sum_{1 \leq k \leq N^i} \sum_{1 \leq s \leq S} W^{i}_s(\tau^{i}_k)}\\
    \tilde{\beta_0} &= \frac{\text{\# medical visits}}{\text{\# medical visits occurring before events}}= \frac{S \times \sum_{1 \leq i \leq N}N^i}{\sum_{1 \leq i \leq N}\sum_{1 \leq k \leq N^i} \sum_{1 \leq s \leq S} (1-W^{i}_s(\tau^{i}_k))}\\
    \beta_1 &= \frac{\tilde{\beta_1}}{\tilde{\beta_1}+\tilde{\beta_0}} \\
    \beta_0 &= \frac{\tilde{\beta_0}}{\tilde{\beta_1}+\tilde{\beta_0}}, \\
\end{align*}
where \#$x$ denotes the cardinal of $x$.

$\mathcal{L}_{2}$ and $\mathcal{L}_{3}$ control for the error over the uncensored samples. $\mathcal{L}_{2}$ penalizes low values of $\tilde{W}$ at the exact time of relapse: 

\[ \mathcal{L}_{2}(Y^i, W^i, \tilde{W}^i) =  \frac{1}{\sum_{1 \leq s \leq S} \sum_{1 \leq i \leq N} Y_s^i} \sum_{1 \leq s \leq S} \sum_{1 \leq i \leq N} \Bigl( Y_s^i \times (1- \tilde{W}^{i}_s(\tau^{i}_{\mu}))^2 \Bigl), \]
\[ \text{ where } \mu = \min \{r | W_s^i(\tau_{r}^i) = 1\} \text{ is the true date of relapse.}\]

$\mathcal{L}_{3}$ penalizes high values of $\tilde{W}$ at the time of the last medical visit before relapse.

\[ \mathcal{L}_{3}(Y^i, W^i, \tilde{W}^i) =  \frac{1}{\sum_{1 \leq s \leq S} \sum_{1 \leq i \leq N} Y_s^i} \sum_{1 \leq s \leq S} \sum_{1 \leq i \leq N} \Bigl( Y_s^i \times (\tilde{W}^{i}_s(\tau^{i}_{\mu - 1 }))^2 \Bigl),\]
\[ \text{ where } \mu = \min \{r | W_s^i(\tau_{r}^i) = 1\} \text{ is the true date of relapse.}\]

$\mathcal{L}_{4}$ controls for the error over the censored samples, by penalizing high values of $\tilde{W}$ at the time of the last medical visit of the sequence.

\[ \mathcal{L}_{4}(Y^i, W^i, \tilde{W}^i) =  \frac{1}{\sum_{1 \leq s \leq S} \sum_{1 \leq i \leq N} (1-Y_s^i)} \sum_{1 \leq s \leq S} \sum_{1 \leq i \leq N} \Bigl( (1-Y_s^i) \times (\tilde{W}^{i}_s(\tau^{i}_{N^i}))^2 \Bigl).\]

The total survival loss function is computed as $\mathcal{L}_{Total} = \alpha_1\mathcal{L}_{1}+\alpha_2\mathcal{L}_{2}+\alpha_3\mathcal{L}_{3}+\alpha_4\mathcal{L}_{4}$, where $\alpha_1, ...,\alpha_4 \in \mathbb{R^+}$.
We used the parameters $\alpha_1,...,\alpha_4$ to constraint all components of the loss function to be of the same order of magnitude. To fix them, we initialized the network and ran one forward pass on the train dataset. We observed the values of $\mathcal{L}_1, \mathcal{L}_2, \mathcal{L}_3$ and $\mathcal{L}_4$ and fixed $\alpha_1$, $\alpha_2$, $\alpha_3$ and $\alpha_4$ accordingly. We used $\alpha_1 = 10$, $\alpha_2 = 1$, $\alpha_3 = 1$, and $\alpha_4 = 1$.

\section{SNDS data preprocessing}\label{apd:snds_data}

In France, all the medical and administrative information relating to the reimbursement of individuals for healthcare expenses are collected and aggregated in the National Health Data System (SNDS) database \citep{tuppin_value_2017}. The longitudinal medical sequences for Cohort A and Cohort B were retrieved from the French administrative claims dataset (\textit{Système National des Données de Santé}, SNDS).
The SNDS contains all individual medical claims in France, including dispensed drugs with date of delivery, laboratory tests, outpatient medical care, and diagnosis and treatments received in hospital (public or private), either for an inpatient stay or for an ambulatory care. The database also contains medical information on the presence of any serious and costly long-term disease giving entitlement to 100\% health insurance coverage (diagnosis code of the disease and disease onset).

In our study, we used procedure, diagnosis and medication codes only. Diagnoses are coded with International Statistical Classification of Diseases and Related Health Problems, Tenth Revision (ICD-10) codes. Procedures are coded with CCAM, a French medical classification of clinical procedures. Medications are primarily coded with the French pharmaceutical categories CIP13; and then mapped to the international ATC (Anatomical Therapeutic Chemical Classification) classification system.

We restricted procedure, diagnosis and medication codes to codes used for diagnostic and treatments of incident BC and BC relapse. We then gathered the medical codes into 47 categories. The list of the 47 categories is displayed in Table \ref{tab:list_medical_codes}. The patients' sequences started at the time of first BC surgery, and were censored at the date of last entry in the Institut Curie records. We combined all consecutive medical visits encompassing the same set of medical codes into a single event starting at the date of the first medical visit among them. 

\begin{table}[hbtp]
\footnotesize
\setlength{\tabcolsep}{3pt}
\floatconts
  {tab:list_medical_codes}
  {\caption{List of medical categories after selection and aggregation of procedure, diagnosis and medication raw codes. Chemotherapy and radiotherapy are defined from both procedure and diagnosis codes.}}
  {\begin{tabular}{p{4.2cm}|p{4cm}|ll}
  \toprule
  \bfseries Procedure & \bfseries Diagnosis & \multicolumn{2}{c}{\bfseries Medication}\\
  \midrule
  Axillary surgery & Breast Cancer & Anastrozole & Gosereline \\
  Breast biopsy & Metastasis & Bevacizumab &  Lapatinib\\
  Breast cytology & Node & BYL719 & Letrozole\\
  Breast imaging & Other cancer & Capecitabine &  Leuproreline\\
  \multicolumn{2}{c|}{Chemotherapy} & Cyclophosphamide &  Melphalan\\
  Lumpectomy & Palliative care & Docetaxel  &  Methotrexate\\
  Lumpectomy/Axillary surgery & Personal history of BC  & Doxorubicine & Mitomycine\\
  Mastectomy & & Epirubicine  &  Paclitaxel \\
  Mastectomy/Axillary surgery & & Eribuline &  Palbociclib\\
  \multicolumn{2}{c|}{Radiotherapy} & Etoposide & Pertuzumab\\
  Node cytology & & Everolimus &  Tamoxifen\\
  Whole body imaging & & Exemestane & Trastuzumab\\
  & & Fluorouracile & Triptoreline\\
  & & Fulvestrant & Vinorelbine\\
  & & Gemcitabine & \\
  \bottomrule
  \end{tabular}}
\end{table}

This study was performed in accordance with institutional and ethical rules concerning research using data from patients. The study was authorized by the French data protection agency (Commission nationale de l’informatique et des libertés—CNIL, under registration numbers DR-2019-006, DR-2020-092, DR-2018-103 and DR-2020-091). No informed consent was required because the data used in the study was de-identified and re-used for research purposes, in accordance with French regulations applicable to the SNDS data.

\section{EDEN implementation details}\label{apd:hyperparameter}

EDEN was trained using Adam optimizer \citep{kingma_adam:_2014} with a learning rate of 0.001 and 500 epochs. At each epoch, we created 10 mini-batches from the train dataset. To account for the large imbalance of the outcome, mini-batches were created by: (1) Splitting the uncensored samples in the training set into 10 batches; (2) Adding at random in each batch the same number of censored samples than uncensored samples. We obtained balanced mini-batches containing roughly 70 samples.

We used the Cohort A validation set to tune the hyper-parameters: (i) embedding size, (ii) hidden size of each of the two T-LSTM networks, (iii) fully connected layer size, (iv) dropout rate for the fully connected layer. We performed random search on the validation set with predefined ranges of the hyper-parameters and 10 epochs. Table \ref{tab:hyperparameter_tuning} displays the predefined ranges for random search, along with the best value found for each hyper-parameter. Results are averaged over 5 independent runs, unless otherwise specified.

\begin{table}[hbtp]
\footnotesize
\setlength{\tabcolsep}{3pt}
\floatconts
  {tab:hyperparameter_tuning}
  {\caption{Predefined ranges for random search and best value of hyper-parameters}}
  {\begin{tabular}{r|c|c}
  & Range & Best value\\
  \midrule
  Embedding size & \{25,50,128,256\} & 50 \\
  Hidden size & \{128,256,512,1024, 2048\} & 128 \\
  Fully connected layer size & \{128,256,512,1024, 2048\} & 1024\\
  Dropout rate & \{0,0.25,0.5,0.75\} & 0.5\\
  \bottomrule
  \end{tabular}}
\end{table}

Once the model trained, we derive $\tilde{Y}^i_s$ as the maximum value of the predicted event rate throughout the period. We inferred that a patient experienced the event $s$ if $\tilde{Y}^i_s>t$ where $t$ is a threshold tuned on the validation set. In that case, we inferred the date of relapse $\tilde{T_s^i}$ as $\tau_{\mu}^i$ where $\mu = \min_{1 \leq r \leq N^i}\big\{r|\tilde{W}_s^i(\tau_{r}^i) \geq t\big\}$.

\section{Baseline implementation}\label{apd:baseline}

\begin{description}
\item[Ad-hoc: ] a set a handmade decision rules based on previously published methods to infer BC relapses from medical claims data \citep{izci_systematic_2020}. We identified BC distant metastatic relapses through the presence of a diagnosis code of metastasis or of a medication code of a treatment specific to metastatic disease (bevacizumab, BYL719, capecitabine, eribuline, etoposide, everolimus, fulvestrant, gemcitabine, lapatinib, melphalan, methotrexate, mitomycine, or palbociclib). Metastatic relapse was dated at the date of the first diagnosis code of metastasis or medication code of a treatment specific to metastatic disease. We identified BC locoregional relapses through the presence of a BC breast surgery code (Lumpectomy, Lumpectomy/Axillary surgery, Mastectomy, or Mastectomy/Axillary surgery) at least one year after the initial BC surgery. Locoregional relapse was dated at the date of the first code BC breast surgery at least one year after the first BC surgery. We identified second cancer relapse through the presence of a diagnosis code of "Other cancer". Second cancer was dated at the date of the first diagnosis code of metastasis.

\item[LSTM: ] a standard many-to-many Long-Short-Term-Memory network,  preceded by an embedding layer and followed by a fully-connected layer. We fixed the embedding size to 50, the hidden layer size to 128, and the fully connected layer size to 1024. We used dropout on the fully connected layer, with a rate of 0.5. We used Binary Cross Entropy (BCE) as the loss function, weighted to account for the over representation of $0$ over $1$ in the event rate function (see the definition of $\mathcal{L}_1$ in Appendix \ref{apd:loss}), and with a learning rate of 0.001.

\item[Dipole \citep{ma_dipole_2017}:] a recently proposed attention-based bi-directional RNN to predict future diagnosis from longitudinal healthcare data \citep{ma_dipole_2017}. To infer the different event rates $W_s$, we extended Dipole to a many-to-many framework, and used $sigmoid$ rather than $softmax$ as the activation function of the fully connected layer.We fixed the embedding size to 50, the hidden layer size to 128, and the fully connected layer size to 1024. We used dropout on the embedding layer, with a rate of 0.5. We used Binary Cross Entropy (BCE) as the loss function, weighted to account for the over representation of $0$ over $1$ in the event rate function (see the definition of $\mathcal{L}_1$ in Appendix \ref{apd:loss}), and with a learning rate of 0.0001.

\item[Timeline \citep{bai_interpretable_2018}:] a novel end-to-end recurrent deep learning approach which uses an attention mechanism to aggregate context information of medical codes and uses time-aware disease specific progression functions to handle irregular timing of events \citep{bai_interpretable_2018}. We extended timeline to a many-to-many framework and used a bi-directional LSTMs as core units. We fixed the embedding size to 50, the hidden layer size to 128, the fully connected layer size to 1024, and the attention mechanism size to 25. We used dropout on the embedding layer, with a rate of 0.5. We used Binary Cross Entropy (BCE) as the loss function, weighted to account for the over representation of $0$ over $1$ in the event rate function, and with a learning rate of 0.001.
\end{description}

\section{Dataset statistics}\label{apd:data_stat}

Table \ref{tab:data_statistic} presents baseline statistics for Cohort A and Cohort B.

\begin{table}
\tiny
\floatconts
  {tab:data_statistic}
  {\caption{Statistics of Cohort A and Cohort B dataset. \#: number; / per. All results are averaged.}}
  {\begin{tabular}{p{3cm}|p{1.3cm}|p{1.3cm}|p{1.3cm}}
  & Cohort A (train+val) & Cohort A (test) &  Cohort B\\
  \midrule
  \# patients & 4,713 & 1,179 & 800\\
  \# events & 116,767 & 28,595 & 14,810\\
  \# events/patient & 24.8 & 24.3 & 18.5\\
  \# of codes/event & 1.4 & 1.4 & 1.8\\
  \midrule
  Censoring rate & 86.4\% & 86.3\% & 91.5\%\\
  Time to censor (days) & 1,814 & 1,800 & 745\\
  \midrule
  Metastasis rate & 6.1\% & 6.1\% & 6.4\%\\
  Time to metastasis (days)  & 1,171 & 1,321 & 679\\
  \midrule
  Locoregional relapse rate & 5.6\% & 5.5\% & 3.0\%\\
  Time to locoregional relapse (days) & 1,337 & 1,411 & 737\\
  \midrule
  $2^{nd}$ cancer rate & 3.8\% & 3.9\% & 0.8\%\\
  Time to $2^{nd}$ cancer (days) & 1,274 & 1,223 & 807\\
  \bottomrule
  \end{tabular}}
\end{table}

\section{Learning curves}\label{apd:loss_curves}

We studied the learning curves dynamic of the different loss terms.
Results are illustrated in Figure \ref{fig:loss_auc_evol}. Recall that EDEN criterion $\mathcal{L}_{total}$ is computed as the sum of four loss functions $\mathcal{L}_1, \mathcal{L}_2, \mathcal{L}_3$ and $\mathcal{L}_4$. The four losses are alternatively optimizing during training, leading to $\mathcal{L}_{total}$ globally decreasing over time, and AUCs increasing over time for the three type of relapse.

The total loss function exhibits peak values where the neural network predicts zero for all time points. The network leaves this trivial solution by making use of the first and second component of the loss function ($\mathcal{L}_1$ and $\mathcal{L}_2$).

We also observe that the AUC curves have different trends for the three types of relapse, with the AUC for metastatic relapse increasing faster than for locoregional and second cancer.

\begin{figure}
\floatconts
  {fig:loss_auc_evol}
  {\vspace*{-10mm} \caption{Learning curves for Cohort A train and validation set on a single run. One epoch stands for an iteration of the learning process through all the batches of the training data.}}
  {\includegraphics[width=\textwidth]{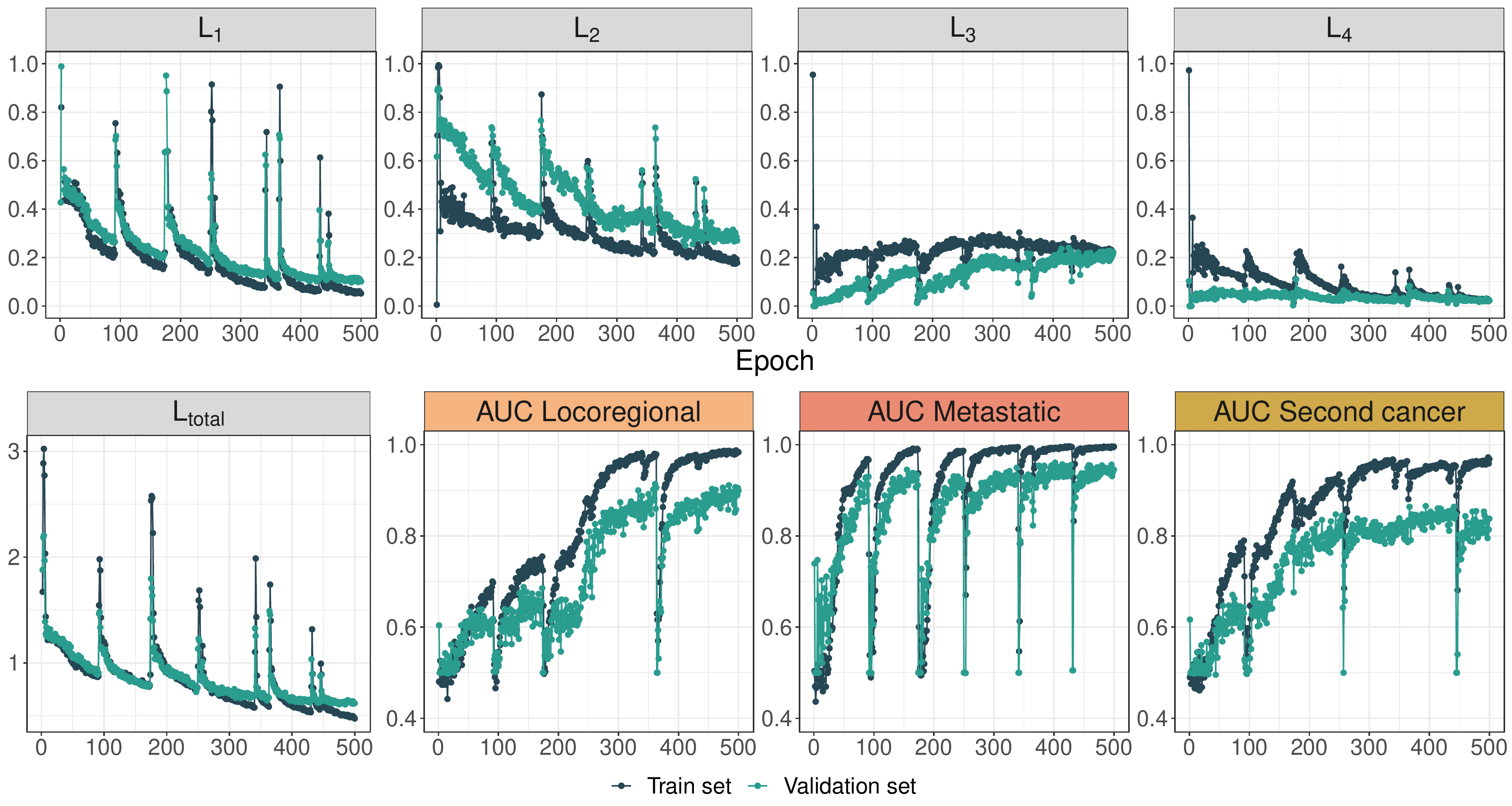}}
\end{figure}

\section{Ablation study}\label{apd:ablation}

\begin{table}
\tiny
\floatconts
  {tab:ablation_study}
  {\caption{Ablation study results on a single run. Models are compared on AUC, accuracy (the \textit{higher}, the better), and mean time interval between the true and the predicted date of relapse $\Delta_T$ (the \textit{lower} in absolute value, the better) for the Cohort A test set. The best scores among models are presented in bold.}}
  {\begin{tabular}{r|c|c|c||c|c|c||c|c|c}
  & \multicolumn{3}{c||}{\color[HTML]{f4a261} Locoregional} & \multicolumn{3}{c||}{\color[HTML]{e76f51} Metastatic} & \multicolumn{3}{c}{\color[HTML]{DFAC2A} Second cancer}\\
  \midrule
  Model & AUC & Acc & $\Delta_T$ & AUC & Acc & $\Delta_T$ & AUC & Acc & $\Delta_T$ \\
  \midrule
  \bfseries LSTM & 94.8 & 94.8 & 89 & 95.6 & 97.1 & 49 & \bfseries 98.1 & 96.9 & 104 \\
  \bfseries T-LSTM & 92.2 & 94.9 & 64 & 97.5 & 96.7 & -7   & 97.8 & 97.2 & 140  \\
  \bfseries Bi-T-LSTM & 96.5 & 97.2 & 38  & \bfseries 98.4 & 98.4 & 37 & 96.4 & 96.4 & 153\\
  \bfseries Bi-T-LSTM - survival output - L1 & 95.9 & 95.4 & 39 & 95.6 & 97.5 & -135 & 92.6 & 97.5 & 289 \\
  
  \bfseries Bi-T-LSTM - survival output - L1 + L2 &  93.7 & 96.3 & -13 & \textbf{98.4} & 98.2 & -165 & 96.4 & 97.6 & \textbf{-77} \\
  
  \bfseries Bi-T-LSTM - survival output - L1 + L2 + L3 &  88.0 & 49.0 & 260 & 97.1 & 78.8 & 201 & 87.4 & 47.1 & 736 \\
  
  \bfseries EDEN & \bfseries 96.9 & \bfseries 97.3 & \bfseries 9    & 98.3 & \bfseries 98.6 & \bfseries -3   & \bfseries 98.1 & \bfseries 98.0   & 230  \\
  \bottomrule
  \end{tabular}}
\end{table}

To determine whether the different components of EDEN architecture and loss improve its performance, we added them one by one from scratch and checked the performance of the so-built models. Indeed, we compared:
\begin{description}
\item[LSTM:] the Long-Short-Term-Memory network used as baseline for the other analyses.
\item[T-LSTM:] the extension of LSTM with time-aware units, as described in \citep{baytas_patient_2017}.
\item[Bi-T-LSTM:] the bidirectional extension of T-LSTM.
\item[Bi-T-LSTM-survival output -L1 :] the Bi-T-LSTM with the proposed survival formulation for the output (meaning we used the chain rule calculation to derive $\tilde{W}$ from $\tilde{h}$) trained with Binary Cross Entropy ($\mathcal{L}_1$);
\item[Bi-T-LSTM-survival output -L1  + L2 :] the Bi-T-LSTM with the proposed survival formulation for the output (meaning we used the chain rule calculation to derive $\tilde{W}$ from $\tilde{h}$) trained with $\mathcal{L}_1$ and $\mathcal{L}_2$;
\item[Bi-T-LSTM-survival output -L1  + L2 + L3 :] the Bi-T-LSTM with the proposed survival formulation for the output (meaning we used the chain rule calculation to derive $\tilde{W}$ from $\tilde{h}$) trained with $\mathcal{L}_1$, $\mathcal{L}_2$ and $\mathcal{L}_3$;
\item[EDEN:] our model; a bi-T-LSTM with the proposed survival formulation for the output and the proposed loss function.
\end{description}

Results are presented in Table \ref{tab:ablation_study}. We see that EDEN outperforms all compared models for all type of relapse. We may conclude that (i) the time-aware unit, (ii) the bidirectional architecture, (iii) the survival formulation for the output; and (iv) the custom loss function, altogether, contribute to the final prediction.

\section{Results interpretation}\label{apd:results_interp}

We examined the impact of each medical code on the predicted event rate function $\tilde{W}_s^i$ to interpret the network results. That is, for each medical code, we averaged the gap in the event rate function $\tilde{W}_s^i(\tau^i_{r+1}) -\tilde{W}_s^i(\tau^i_{r-1})$ for all medical visits $r$ during which the given medical code was reported. Results are presented in Figure \ref{fig_logit_gap.pdf}. 

\begin{figure*}
\floatconts
  {fig:logit_gap}
  {\caption{Average gap in the predicted event rate function for each medical code and each type of BC relapse on a single run. To ensure readability, we displayed only the top 20 medical codes for each type of relapse.}}
  {\includegraphics[width=0.9\textwidth]{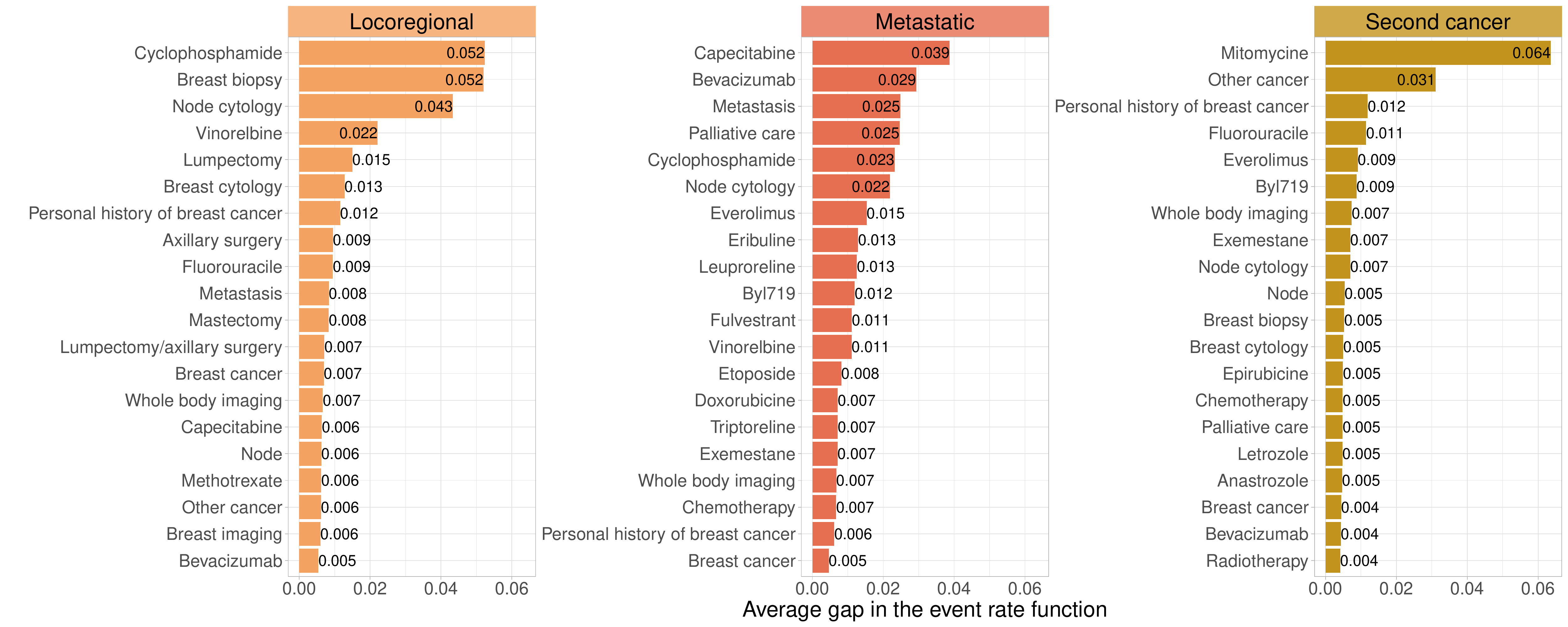}}
\end{figure*}

%The results are presented in Figure \ref{fig_logit_gap.pdf}. They were in accordance with clinical practices,  with codes of molecules approved only in the metastatic setting associated with an increase in the event rate function for metastatic relapse. The recording of local BC diagnostic (breast biopsy and cytology) or surgical procedures (lumpectomy) was associated with an increase in the event rate function of locoregional relapse.

\section{Survival curves}\label{apd:survival_curves}

We drew Kaplan-Meier estimated curves of actual \textit{versus} predicted survival. Kaplan-Meier method estimates the survival rate $S_s^i(t)$, \textit{i.e.} the probability of surviving a given length of time: $S_s^i(t) = Pr(T_s^i > t) = 1 - W_s^i(t)$.

\begin{figure*}
\floatconts
  {fig:loss_survival_curve}
  {\caption{Kaplan-Meier survival curves for predicted and true survival data on the Cohort A test set on a single run.}}
  {\includegraphics[width=0.9\textwidth]{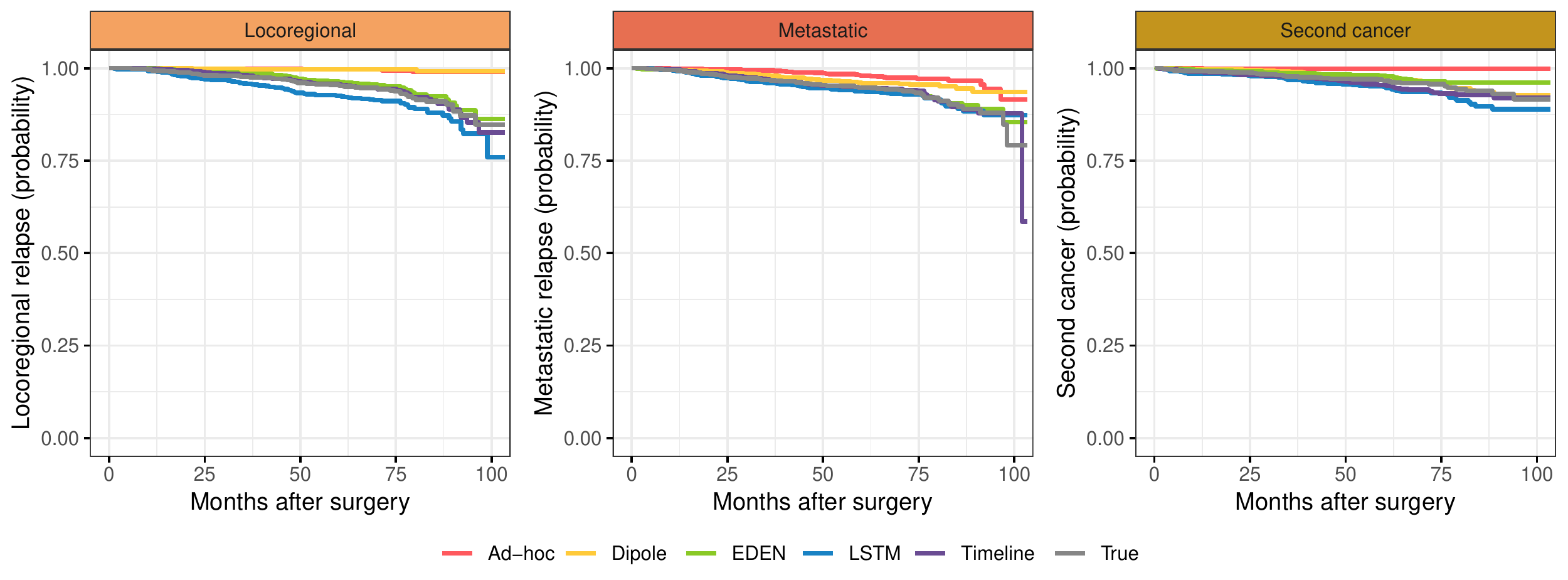}}
\end{figure*}

The Kaplan-Meier survival curves for both true and predicted relapse for all models are displayed in
Figure \ref{fig:loss_survival_curve}. The curves obtained from EDEN prediction almost overlapped with the true survival curves, suggesting that EDEN was able to capture survival properties at the population level.

%\vspace{-15cm}
\end{document}